\documentclass[12pt,draftcls,onecolumn]{IEEEtran}
\IEEEoverridecommandlockouts

\usepackage{enumerate}
\usepackage[english]{babel}
\usepackage[utf8]{inputenc}
\usepackage{algorithm}
\usepackage[noend]{algpseudocode}
\usepackage{amsmath}
\usepackage{amssymb}
\usepackage{amsthm}
\usepackage{graphicx}
\usepackage{hyperref}
\usepackage{graphics}
\usepackage{subcaption}
\usepackage{array}
\usepackage{cite}

\newcolumntype{C}[1]{>{\centering\arraybackslash}m{#1}}

\title{Attention Actor-Critic algorithm for Multi-Agent Constrained Co-operative Reinforcement Learning \footnote{Equal contribution by the first three authors. A version of this paper has been accepted for publication as an extended abstract in the Proceedings of the 20th International Conference on Autonomous Agents and Multiagent Systems
(AAMAS 2021).}}

\author{P. Parnika$^{*,1}$, Raghuram Bharadwaj Diddigi$^{*,2}$, Sai Koti Reddy Danda$^{*,3}$ and  Shalabh Bhatnagar$^{2}$

$^1$ Mindtree Ltd.\\
$^2$ Department of Computer Science and Automation, IISc Bangalore, India.\\
$^3$ IBM Research, Bangalore, India\\
parnika.ajay@mindtree.com, \{raghub,shalabh\}@iisc.ac.in,
saikotireddy@in.ibm.com
}

\begin{document}

\maketitle
\thispagestyle{empty}
\pagestyle{empty}

\begin{abstract}
In this work, we consider the problem of computing optimal actions for Reinforcement Learning (RL) agents in a co-operative setting, where the objective is to optimize a common goal. However, in many real-life applications, in addition to optimizing the goal, the agents are required to satisfy certain constraints specified on their actions. Under this setting, the objective of the agents is to not only learn the actions that optimize the common objective but also meet the specified constraints. In recent times, the Actor-Critic algorithm with an attention mechanism has been successfully applied to obtain optimal actions for RL agents in multi-agent environments. In this work, we extend this algorithm to the constrained multi-agent RL setting. The idea here is that optimizing the common goal and satisfying the constraints may require different modes of attention. By incorporating different attention modes, the agents can select useful information required for optimizing the objective and satisfying the constraints separately, thereby yielding better actions. Through experiments on benchmark multi-agent environments, we show the effectiveness of our proposed algorithm.
\end{abstract}

\section{Introduction}
\label{intro}
In a multi-agent co-operative RL setting \cite{marlsurvey}, multiple agents are working towards a common goal in a common environment. All the agents receive the same cost (or reward) depending on the actions of all the agents and the objective is to minimize (or maximize) the expected total discounted cost (or reward) \cite{sutton2018reinforcement}. However in many practical situations, one often encounters constraints that restrict the choice of actions that can be taken by these agents. In the constrained RL setting \cite{altman2005zero}, these constraints can also be specified via certain expected total discounted costs. In such scenarios, the agents have to learn actions that not only minimize the expected total discounted cost but also respect the constraints. 

One approach to satisfy the constraints is to construct a modified cost as a linear combination of the original cost and the constraint costs. However, the weights to be associated with the costs are not known upfront and need to be learned in a trial-and-error fashion. This problem becomes compounded when multiple constraints are specified. We alleviate this problem by considering the Lagrangian formulation of the problem and training dual Lagrange parameters that act as weights for the constraint costs. 
\begin{table}[h!]
\centering
\begin{tabular}{|C{4cm} | C{8cm} |}
     \hline
     \textbf{References} & \textbf{Features}  \\ \hline
     \cite{lowe2017multi,foerster2017counterfactual,chen2019new,nguyen2020deep,oroojlooyjadid2019review} & Deep RL algorithms for multi-agent setting. Attention mechanism not considered. \\ \hline
     \cite{pmlr-v97-iqbal19a,mao2018modelling, jiang2018learning} & Deep RL algorithms with Attention for multi-agent setting. Constrained setting not considered. \\ \hline
     \cite{borkar2005actor,bhatnagar2012online,bhatnagar2010actor} & RL algorithms for single-agent constrained setting. Multi-agent constrained setting not considered. \\ \hline
     \cite{achiam2017constrained,tessler2018reward,liang2018accelerated} & Deep RL algorithms for single-agent constrained setting. Multi-agent constrained setting not considered.  \\ \hline
     \cite{boutilier2016budget,zhang2016decision,agrawal2016scalable,dolgov2011resource,shalev2016safe,fowler2018constrained, reddy2019risk, diddigi2019actorAamas} & RL algorithms for multi-agent constrained setting. Attention mechanism not considered. \\ \hline
     \textbf{Our Work} & Deep RL algorithm with Attention mechanism for multi-agent Constrained setting. \\ \hline
\end{tabular}
\vspace{0.3 cm}
\caption{Comparison with other works in the Literature}
\label{rel-works}
\end{table}
\begin{figure*}[ht]
\centering
\includegraphics*[scale=0.65]{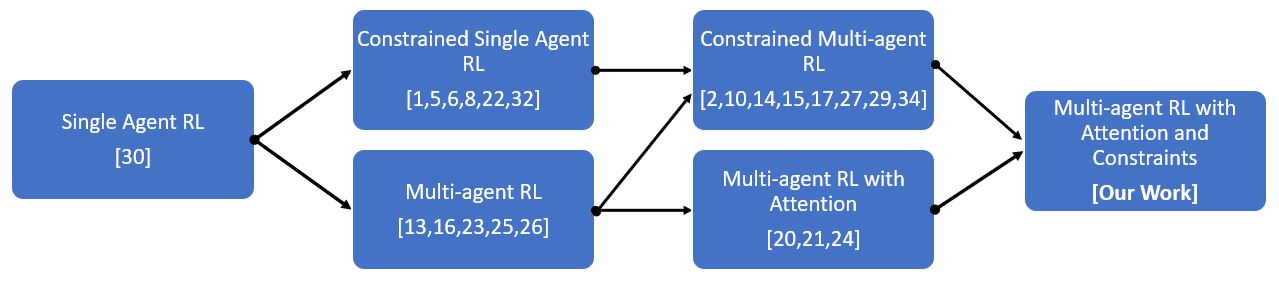}
\caption{Evolution of paradigms in literature}
\label{seq_flow_diag}
\end{figure*}

Single-agent RL algorithms for the constrained RL settings have been proposed under various cost criteria like average cost in \cite{borkar2005actor,bhatnagar2012online} and discounted costs in \cite{bhatnagar2010actor,achiam2017constrained,tessler2018reward,liang2018accelerated}. Constraints in a multi-agent setting can appear in multiple ways. Under budget constraints \cite{boutilier2016budget}, every joint policy is associated with a cost. The objective of the agents here is to compute a joint optimal policy that maximizes the value, respecting the budget constraints. Under resource/task constraints \cite{agrawal2016scalable,dolgov2011resource,fowler2018constrained}, the optimal policy is the one that not only maximizes the value but also optimally allocates the resources to the agents. Under safety constraints \cite{zhang2016decision,shalev2016safe,reddy2019risk}, each policy is associated with a safety value, and the objective of the agents is to compute optimal policies that meet the safety constraints. Finally, in \cite{diddigi2019actorAamas}, similar to the model we consider in this paper, the constraints are specified as expected discounted cost which are required to be less than a prescribed threshold value. 

Actor-Critic algorithms \cite{sutton2018reinforcement} are a popular class of RL algorithms that are used by an agent to obtain an optimal policy. In this paradigm, `Actor' computes the policy and `Critic' provides feedback on the policy computed by the `Actor'. Based on this feedback, `Actor' improves the policy. This process is repeated until an optimal policy is obtained. The Actor-Critic paradigm for multi-agent settings can be extended in three ways \cite{oroojlooyjadid2019review}. All the agents can independently (without co-operation and communication) run the Actor-Critic algorithm. This setting of agents is known as `Independent Learners' \cite{tampuu2017multiagent} and it suffers from the problem of non-stationarity \cite{marlsurvey}. Another setting known as `Joint Action Learners' assumes the existence of a central controller which computes the optimal policy of all the agents and communicates the actions to the agents. This setting suffers from scalability problems as the state and action spaces for the central controller increase exponentially as the number of the agents increases. Finally, a paradigm that mitigates the problem of scalability and non-stationarity known as `centralized learning and decentralized execution' has become popular in recent times \cite{lowe2017multi,foerster2017counterfactual,chen2019new,nguyen2020deep}. The main idea here is to use a centralized critic during the training and decentralized actors that learn actions independently. In these algorithms, however, information of all the agents are given equal importance (or weights) while computing the optimal policy.  

The attention mechanism allows an agent to selectively pay attention to those agents whose information is crucial in the computation of its policy. In \cite{pmlr-v97-iqbal19a}, attention actor-critic algorithms have been proposed that make use of the attention mechanism in the learning of `critic'. In \cite{mao2018modelling}, attention mechanism has been used to model the policies of teammates. The attention mechanism for learning communication among the agents has been proposed in \cite{jiang2018learning}. In this work, the authors propose an attentional communication model ATOC that provides an effective mechanism for communication among agents resulting in better decision making. We illustrate the advantages of the attention mechanism through the following example. Consider a Smart Grid setup \cite{saad2012game} where a network of microgrids are employed, whose objective is to provide power to its dedicated customers. These microgrids are equipped with renewable energy generation sources (like solar panels, wind turbines) and limited storage battery device to store their power. At every time instant, each microgrid has to make intelligent decisions like the number of units of power to store in its battery, the number of units of power to buy (or sell) from (to) other microgrids to maximize its profits. For any given microgrid, the state information of its neighboring microgrids is more important than those at a far distance from it. Through the attention mechanism, a microgrid can dynamically select these neighboring microgrids, instead of attending to all the microgrids equally. This results in better decision making and hence better profits for the microgrids. 


We believe that the attention mechanism is particularly important in the constrained multi-agent setting. We explain its importance through the two following examples. 
\begin{enumerate}
    \item Consider a warehouse where multiple robots are deployed. The objective of the robots is to pick up the goods from a target position with a constraint that the expected number of collisions among the robots is less than a predefined threshold. The important information for a robot for collecting goods is the position of goods whereas the relative distance between the robots is crucial information to avoid collisions. Hence having attention mechanisms separately for learning optimal actions to collect the goods and avoid collisions will be natural in this setting.
    \item In the smart grid setting considered, let's say that we impose a constraint that the expected demand-supply deficit should be maintained at a certain level (to ensure the stability of the grid). The relevant information for maximizing the profits and maintaining the stability for a microgrid can be different. The attention mechanism enables the microgrid to attend to the relevant information for these two tasks separately. 
\end{enumerate}
Moreover, the attention mechanism for multi-agent constrained setting finds its applications in numerous settings like self-driving cars \cite{shalev2016safe} to ensure safety constraints, supply chain optimization \cite{giannoccaro2002inventory} to ensure the resource constraints. In our work, we propose an attention-based Actor-Critic algorithm for solving the problem of multi-agent constrained Reinforcement Learning (RL). While the attention mechanism and constrained RL settings have been studied extensively in the literature, the use of two separate attention mechanisms for computing the policy and satisfying the constraints has not been considered previously. We believe that this architecture is very important as optimizing the common goal and satisfying the constraints require different modes of attention. We show through our analysis of attention weights that using multiple attentive critics can benefit and yield much better results on complicated real-world applications. A brief overview of the comparison of our work with other works in the literature is provided in Table \ref{rel-works} and the evolution diagram of these paradigms/themes is shown in Figure \ref{seq_flow_diag}. The main contributions of our work are the following:
\begin{itemize}
    \item We propose an Actor-Critic algorithm for computing the optimal actions for agents in a constrained co-operative multi-agent setting that makes use of the attention mechanism.
    \item We analyze and discuss the performance of our algorithm on constrained versions of standard multi-agent RL environments. 
    \item We provide a detailed analysis of the attention mechanism learned by the agents in our experiments (Section \ref{attn-disc}).
\end{itemize}
The rest of the paper is organized as follows. In Section \ref{model}, we describe the multi-agent constrained co-operative setting considered in the paper. In Section \ref{algo}, we propose our multi-agent attention mechanism-based constrained Actor-Critic algorithm. In Sections \ref{exp} and \ref{ab-studies}, we present the performance of our algorithm on multi-agent environments and discuss the results. Concluding remarks are given in Section \ref{concl}.


\section{Model}
\label{model}
We now discuss the constrained co-operative multi-agent setting described in \cite{diddigi2019actor}. 
It is described by tuple $<n,S,A,T,k,c_1,\ldots,c_m,\gamma>$. Here, $n$ denotes the number of agents in the environment. $S = S_1 \times S_2 \times \ldots S_n$ is the joint state space and $s \in S = (s_1 \ldots s_n)$ is the joint state with $s_i \in S_i$ being the state of the agent $i$. Similarly, $A = A_1 \times,\ldots,\times A_n$ denotes the joint action space where $a \in A = (a_1,\ldots,a_n)$ is the joint action and $a_i \in A_i$ being the action of agent $i$. Each agent only observes its own state and chooses its action based on it. Let $T$ be the probability transition matrix where $T(s'|s,a)$ denotes the probability of next state being $s'$ when joint action $a$ is taken in joint state $s$. 
Single-stage cost function ($k$) is the cost incurred when joint action $a$ is taken in state $s$.
Moreover, $c_1,\ldots,c_m$ denote the single-stage cost functions for the constraints. Note that both the main cost function ($k$) and constraint costs  ($c_1,\ldots,c_m$) depend on the joint action of the agents. Finally, $\gamma$ denotes the discount factor. Let $\pi_i:S_i \xrightarrow{} \Delta(A_i)$ denote the policy of agent $i$, where for a given state of agent $i$, $\pi_i(s_i)$ is a probability distribution over its actions. We now define the total discounted cost ($J$) for a joint policy $\pi = (\pi_1,\ldots,\pi_n)$ as follows: 
\begin{align}
\label{main-cost}
    J(\pi) = E \Big{[}\displaystyle \sum_{t = 0}^{\tau} \gamma^t k(s_t,\pi(s_t)) \Big{]},
\end{align}
where $E(.)$ is the expectation over entire trajectory of states with initial state $s_0 \sim d_0$,  where $d_0$ is a probability distribution over states, $\tau$ is a finite stopping time and $s_t$ is the joint state at time $t$. 
The $m$ constraints on the system are defined as follows:
\begin{align}
\label{cons-cost}
    E \Big{[} \displaystyle \sum_{t=0}^{\tau} \gamma^{t} c_j(s_t,\pi(s_t)) \Big{]} \leq \alpha_j, ~ \forall j \in {1,\ldots,m},
\end{align}
where $\alpha_1,\ldots,\alpha_m$ are pre-specified thresholds.

The objective of the agents in the multi-agent constrained co-operative RL setting is to compute a joint policy $\pi^* = (\pi_1^*,\ldots,\pi_n^*)$ that 
\begin{align}
\label{comb-cost}
&\min_{\pi \in \Pi}\,\,\, J(\pi) = \,\,E \Big{[}\displaystyle \sum_{t = 0}^{\tau} \gamma^t k(s_t,\pi(s_t)) \Big{]} \\
\nonumber 
    &\text{s.t}\,\,\,  ~ E \Big{[} \displaystyle \sum_{t=0}^{\tau} \gamma^{t} c_j(s_t,\pi(s_t)) \Big{]} \leq \alpha_j, ~ \forall j \in {1,\ldots,m},
\end{align}
where $\Pi$ is set of all joint policies. 
The constrained problem \eqref{comb-cost} can be relaxed using the Lagrangian formulation \cite{borkar2005actor,bhatnagar2010actor} as follows:
\begin{align}
    \label{lag-cost}
    L(\pi,\lambda) = E \Big{[}\displaystyle \sum_{t = 0}^{\tau} \gamma^t \big{(} k(s_t,\pi(s_t)) + \displaystyle \sum_{j=1}^{m} \lambda_j c_j(s_t,\pi(s_t)) \big{)} \Big{]} - \displaystyle \sum_{j=1}^{m} \lambda_j\alpha_j,
\end{align}
where $\lambda = (\lambda_1,\ldots,\lambda_m)$ is the vector of Lagrange parameters associated with the $m$ constraints. 

From the theory of duality in optimisation (Chapter 5 of \cite{boyd2004convex}), it is clear that the optimal policy $\pi^*$ and Lagrange parameters $\lambda^*$ satisfy the following:

\begin{align}
    \label{dual}
    L(\pi^*,\lambda^*) = \displaystyle \max_{\lambda > 0} \min_{\pi \in \Pi} L(\pi,\lambda).
\end{align}

The theory of two time-scale stochastic approximation \cite{borkar1997stochastic} allows us to iteratively learn the Lagrange parameters and policy. The main idea is to perform gradient descent on the objective \eqref{lag-cost} in the space of policy parameters on the faster timescale and gradient ascent on \eqref{lag-cost} in the Lagrange parameters on the slower timescale \cite{borkar2009stochastic}. The complete details of how we achieve this is described in the next section. 
\section{Proposed Algorithm}
\label{algo}
\textbf{Attention mechanism \cite{xu2015show,mao2018modelling}:} In general, the attention mechanism works as follows. It takes  source vectors $v = (v_1,\ldots,v_n)$ and a target vector $T$ as inputs and outputs a context vector $C$. The attention mechanism first computes attention weights $(w_1,\ldots,w_n)$, where $w_i, ~ 1 \leq i \leq n$ represents the importance of $v_i$. The attention weight $w_i$ is computed from a given function $f(T,v)$ as follows:
\begin{align}
\label{attention-1}
    w_i = softmax(f(T,v)) = \frac{exp(f(T,v_i))}{\sum_{j=1}^{n} exp(f(T,v_j))}.
\end{align}
Finally, the context vector is computed as:
\begin{align}
\label{attention-2}
    C = \sum_{j=1}^{n} w_j v_j.
\end{align}
As it can be seen from \eqref{attention-1} that $\sum_{j=1}^{n} w_j = 1$, the attention mechanism can thought of a computation that adaptively learns the distribution over input vector that accurately represent the context of the problem.

We extend the attention mechanism proposed in the context of the multi-agent RL setting \cite{pmlr-v97-iqbal19a} to the constrained setting. The details of the proposed attention mechanism are as follows. Each agent $i$ maintains a total of $m+1$ critics which use attention. Let’s denote these as the cost critic $Q_\psi$ (associated with the main cost function) and $m$ penalty critics $Q_{\eta 1},\ldots, Q_{\eta m}$ (associated with the $m$ constraints). 
The intuition here is, by having multiple critics with different attentions, each critic is especially able to attend to that information which is crucial in solving its objective. The way this information is utilized for attention is by encoding state and state-action information of all agents where the embedding function is a single layer perceptron.

These encodings are passed to another embedding function, also a single layer perceptron, to create keys ($K$), values ($V$), and selectors/queries ($q$) \cite{pmlr-v97-iqbal19a}. The keys $K_j$ and values $V_j$ represent state-action encodings of all agents $j \neq i$ while queries $q_i$ are state encodings of the agent $i$. Now, the attention weights $w_j$ are computed as a function of queries and keys as follows:
\begin{align}
    w_j = softmax\Big{(} \frac{q_iK_j^T}{\sqrt{d_k}}\Big{)},
\end{align}
where $d_k$ is the size of the keys. 
Finally, critic $Q$ of agent $i$ (denoted by $Q^i$) is obtained as follows (for notation convenience, we drop the subscripts from the critics of agent $i$ as all $m+1$ critics use similar architecture):
\begin{align}
    Q^i = f_i(g_i(o_i,a_i),x_i),
\end{align}
where $f_i$ is a multi-layer perceptron with two layers, $g_i$ is an embedding function for agent $i$, and $x_i = \displaystyle \sum_{j\neq i}w_jV_j$ is the contribution of other agents to agent $i$. 

We now discuss our proposed algorithm `MACAAC' (Algorithm \ref{C-AC}). We train the algorithm in $\mu$ parallel environments to improve the sample efficiency and reduce the variance of updates. At each time step of an episode, every agent samples an action from its current policy based on its observations $o_i$ and obtains common single-stage cost $k$, single-stage penalties $c_1,\ldots,c_m$ and next state as shown in steps 19 of Algorithm \ref{C-AC}. The Lagrangian cost is calculated as shown in the step 20. This information is then stored in the replay buffer $D$. The `Critic' and `Actor' parameters are updated after every $U$ steps. This is done as follows.  First a minibatch `B' is sampled independently from the replay buffer. For each sample from the minibatch `B', the critic parameters are updated (Step 27 of Algorithm \ref{C-AC}) by performing gradient descent on the MSE loss given by \cite{pmlr-v97-iqbal19a}:
\begin{align}\label{criticUpdates}
    \displaystyle \sum_{i=1}^{n}E[(Q^i(o,a) - y_i)^2],
\end{align}
where $y_i = r + \gamma E[Q^i(o^*,a^*) - \alpha \log(\pi_\theta(a_i^*|o_i^*))]$, $o^*,a^*$ are the joint next state and actions of agents and $\alpha$ is known as the temperature coefficient  \cite{pmlr-v97-iqbal19a} that is used to control the stochastic nature of the policy. The parameters of cost critic $\psi$ are updated by performing gradient descent on \eqref{criticUpdates} with $r$ defined as in Line 20 of Algorithm \ref{C-AC} (Lagrangian cost) and the parameters of penalty critic $j$, $\eta_j$, are updated by performing gradient descent on \eqref{criticUpdates} with $r$ as $c_j$ (constraint cost). Moreover, all the agents $i$ share the same parameters ($\psi,\eta_1,\ldots,\eta_m$) of critics.

The `UpdateActors' step is performed as follows. The policy parameters of each agent ($\theta_i$) are updated by performing gradient descent using the gradient function given by \cite{pmlr-v97-iqbal19a}:
\begin{align}\label{actorupdates}
    E[\nabla_{\theta_i}\log(\pi_{\theta_i}(a_i|o_i))(-\alpha \log(\pi_{\theta_i}(a_i|o_i)) + Q_\psi^i(o,a) - b(o,a_{-i}))],
\end{align}
where $b$ is a baseline function that is independent of actions of agent $i$ ($a_{-i}$ represents the actions of all agents except $i$). Note that in equations \eqref{criticUpdates} and \eqref{actorupdates}, an entropy term is added that facilitates stochastic policies \cite{haarnoja2018soft}.

Finally, the Lagrange parameters $\lambda_j, ~ j = 1,\ldots,m$ are updated by performing gradient ascent on the Lagrangian $L$ (eq. \ref{lag-cost}) as shown in the step 30. An important point to note here is that the critic and actor updates (steps 27 and 28) are performed on a faster time-scale compared to the Lagrange parameter updates (steps 29-30). As a result, the critic and actor perceive Lagrange parameters as constants in their updates, thereby ensuring the convergence of the algorithm \cite[Chapter 6]{borkar2009stochastic}.  
\begin{algorithm}
\caption{Multi-Agent Constrained Attention Actor-Critic (MACAAC)}
\label{C-AC}
\begin{algorithmic}[1]
    \State $E \xleftarrow[]{}$ Maximum number of episodes.
    \State $L \xleftarrow[]{}$ Length of an episode.
    \State $U \xleftarrow[]{}$ Steps per update.
    \State $\theta_i \xleftarrow[]{}$ policy parameters of the agent $i$, $i = 1,\ldots,n$.
    \State \textbf{UpdateCritic:} Subroutine to update the critic parameters.
    \State \textbf{UpdateActors:} Subroutine to update the policy parameters of all the agents. 
    \State $Q_{\eta_{j}} \xleftarrow[]{}$ Q-value of constrained cost associated with constraint $j, ~ j = 1,\ldots,m$.
    \State $\beta_t \xleftarrow[]{}$ Slower timescale step-size at time step $t$.
    \State Initialize Lagrange parameters $\lambda_1,\ldots,\lambda_m$.
    \State Create $\mu$ parallel environments.
    \State Initialise replay buffer, D.
    \State $u \xleftarrow[]{} 0 $
    \For{$ep = 1,2,\ldots, E$}
    \State Obtain initial observations $o_i^e$ for all agents $i$ in each \\ \hspace{0.4 cm} environment $e$
    \For{$t$ = $1,2,\ldots,L$}
    \State Obtain actions $a_i^e \sim \pi_{\theta_i}(.|o_i^e)$, $\forall i = 1,\ldots,n$, \\ \hspace{1 cm}$\forall e = 1,\ldots,\mu$
    \State Execute actions and get $(o^{*,e}_{i},k^e,c_{1}^e,c_{2}^e,\ldots,c_{m}^e)$ ~ $\forall i,e$
    \State Let $r^e = k^e + \displaystyle \sum_{j=1}^{m}\lambda_j c_{j}^e$, ~ $\forall e$
    \State Store $(o^{e}_{i},a_i^e,r^e,c_{1}^e,c_{2}^e,\ldots,c_{m}^e,o^{*,e}_{i})$, $\forall i,e$ in $D$
    \State $o_i^e = o_i^{*,e}$, $\forall i,e$
    \State $u += \mu$
    \If{($u \%$ U) < $\mu$}
    \State Sample minibatch (B) from D
    \State Get next actions $a^{'}_1,\ldots,a^{'}_n$
    \State \textbf{UpdateCritic}(B, $a^{'}_1,\ldots,a^{'}_n$)
    \State \textbf{UpdateActors}(B)
    \For{$j = 1,\ldots,m$}
    \State $\lambda_j \xleftarrow[]{} \max(0,\lambda_j + \beta_t(Q_{\eta_ {j}} - \alpha_j))$
    \EndFor
    \EndIf
    \EndFor
    \EndFor
	\end{algorithmic}
\end{algorithm}

\section{Experiments and Results}
\label{exp}
In this section, we describe the performance of our proposed Algorithm `MACAAC' on two multi-agent environments and analyze the results. The first environment is the constrained version of Cooperative Navigation \cite{lowe2017multi} followed by the constrained version of Cooperative Treasure Collection \cite{pmlr-v97-iqbal19a}. The constraint considered in the experiments is the collision between the multiple agents. The agents incur a penalty whenever there is a collision and their objective is to make sure that the expected total penalty is less than a prescribed penalty threshold. To avoid confusion, we refer to the main cost that the agents are minimizing as `cost' and the constrained cost as the `penalty'. For comparison purposes, we also implement the constrained version of MADDPG \cite{lowe2017multi} algorithm, which we refer to as `MADDPG-C'. Moreover, to better analyze the results, we also report the results on an un-constrained version of Multi-agent Attention Actor-Critic \cite{pmlr-v97-iqbal19a} where there is no penalty incurred for collisions among the agents, which we simply refer to as `Unconstrained'. Finally, in section \ref{ab-studies}, we evaluate the performance of `MACAAC' with fixed weights. The neural network architecture and hyper-parameters are kept the same for all three algorithms \footnote{The source codes of our experiments are available at: \url{https://github.com/parnika31/MACAAC_Supplementary} }  
\subsection{Constrained Cooperative Navigation}
\subsubsection{Description of the experiment}
\begin{figure}[ht]
\centering
\fbox{\includegraphics[scale=0.28]{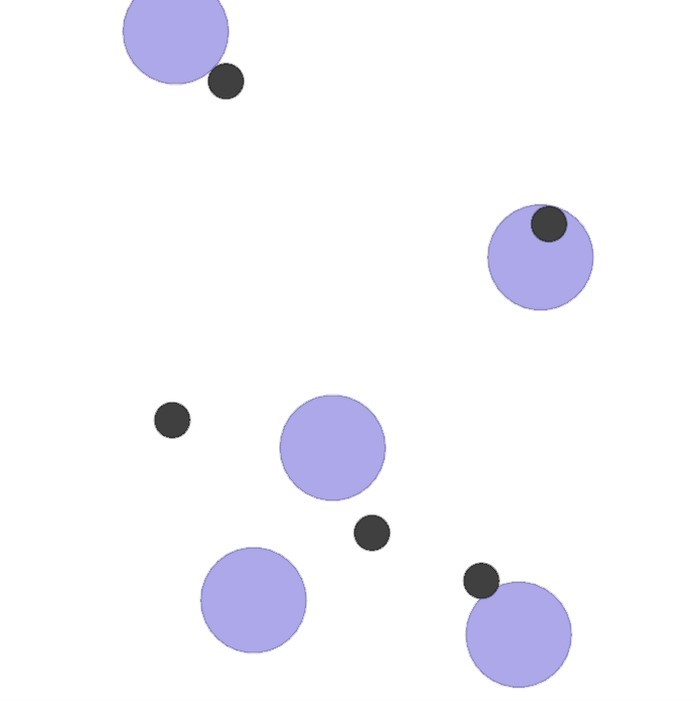}}
\caption{Constrained Cooperative Navigation. The large blue balls are `agents', whose objective is to navigate towards the small black balls which are `targets' without collisions.}
\label{cn-ss}
\end{figure}

In this experiment, there are $5$ agents and $5$ targets that are randomly generated in a continuous environment at the beginning of each episode as shown in the Figure \ref{cn-ss}.
The objective of the agents is to navigate towards the targets in a co-operative manner such that all targets are covered. The length of each episode is $25$ time steps and the single-stage cost at each time step is the sum of the distance to the nearest agent, over all the targets. Therefore, the agents have to learn to navigate towards the targets in such a way that all target positions are covered. However, we include a single-stage penalty of $1$ when there is a collision between the agents (and $0$ otherwise). The penalty threshold ($\alpha$) is set to $3$ in our experiments. This means that the expected total penalty over all the episodes must be less than or equal to $3$. The discount factor is set to $0.99$. In Figure \ref{ccn-plots}, we show the performance of algorithms during the training phase, and in Table \ref{t1}, we report the performance of algorithms during the testing.  

\subsubsection{Discussion}
\begin{figure}[ht]
\centering
        \begin{subfigure}{0.5\textwidth}
              \includegraphics[width=0.95\columnwidth]{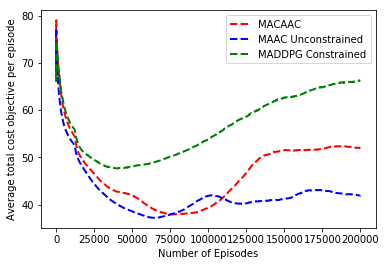}
                \caption{Expected total cost}
                \label{cost-ccn}
        \end{subfigure}%
        \begin{subfigure}{0.5\textwidth}
                \includegraphics[width=0.95\columnwidth]{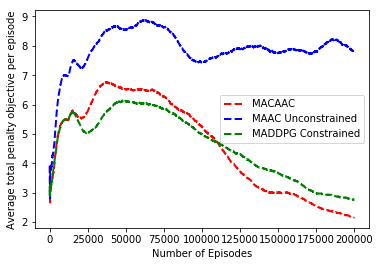}
                \caption{Expected total penalty}
                \label{pen-ccn}
        \end{subfigure}%
        \caption{Performance of Algorithms on Constrained Cooperative Navigation during the training. The average total cost and penalty at each episode $i$ are calculated, by taking mean of total costs and total penalties over 1024 runs, using the policies trained until $i$ episodes.}\label{ccn-plots}
\end{figure}

\begin{itemize}
    \item In Figure \ref{cost-ccn}, we observe that the total cost approaches convergence for all the three algorithms. The `Unconstrained' algorithm achieves the smallest average cost as there is no penalty for collisions in this case. Therefore, the agents can move freely in the continuous space and navigate quickly towards the targets. This can also be observed in Figure \ref{pen-ccn}, where we see that the average penalty of the `Unconstrained' algorithm is the highest. 
    \item In Figure \ref{pen-ccn}, we see that the average penalty comes down as the training progresses for the constrained algorithms (MADDPG-C and our proposed MACAAC), while for the `unconstrained' algorithm it almost remains constant. This is the effect of Lagrange parameters that are learnt in the constrained setting. 
    \item From Table \ref{t1}, we can see that both our proposed algorithm `MACAAC' and `MADDPG-C' satisfies the penalty constraint. However, our algorithm `MACAAC' achieves this average penalty at a lower cost than the `MADDPG-C' algorithm. 
\end{itemize}
\begin{figure}[ht]
\centering
\fbox{\includegraphics[scale=0.28]{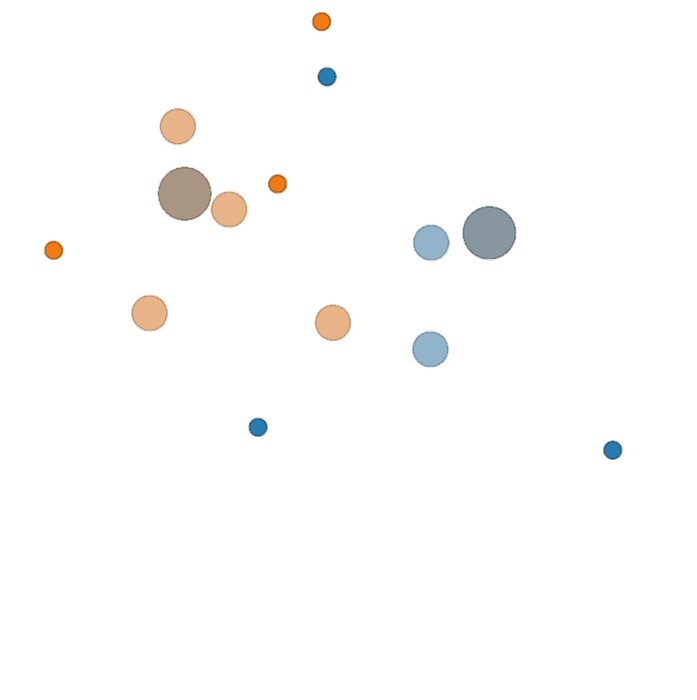}}
\caption{Constrained Cooperative Treasure Collection. The big blue and brown balls are `depositors'. The dark orange and blue colored balls are `treasures' which will be re-spawned after every capture. The rest of the balls are `collectors'. In this figure, the collector agents changed color after capturing the treasures and are moving towards depositors (or banks) of same color to deposit them without collisions.}
\label{ctc-ss}
\end{figure}

\begin{table}[ht]
\renewcommand{\arraystretch}{1.3}
\centering
\begin{tabular}{|c|c|c|}
\hline
Name of the Algorithm                 & \begin{tabular}[c]{@{}c@{}}Average total cost over $10,000$\\ episodes \end{tabular} & \begin{tabular}[c]{@{}c@{}}Average total penalty over $10,000$ \\ episodes\end{tabular} \\ \hline
\textbf{MACAAC} & \textbf{45.79}                                                                                & \textbf{1.87}                                                                                      \\ \hline
MADDPG-C                                & 60.33                                                                                 & 2.52                                                                                      \\ \hline
Unconstrained                         & 37.50                                                                                & 7.02                                                                                      \\ \hline
MACAAC with Fixed Weights                         & 38.73                                                                                & 1.25                                                                                      \\ \hline
\end{tabular}
\vspace*{0.3 cm}
\caption{Performance comparison of algorithms in testing phase on Constrained Cooperative Navigation with penalty threshold $\alpha$= $3$. The average total cost and penalty are calculated by taking mean of total costs and penalties, respectively, over $10,000$ runs using the policies of agents obtained at the end of training.}
\label{t1}
\end{table}
\begin{figure}[ht]
\centering
        \begin{subfigure}{0.3\textwidth}
        \centering
                \includegraphics[width=0.95\columnwidth]{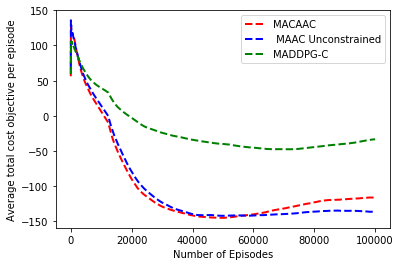}
                \caption{Expected total cost}
                \label{cost-ctc}
        \end{subfigure}%
        \begin{subfigure}{0.3\textwidth}
        \centering
                \includegraphics[width=0.95\columnwidth]{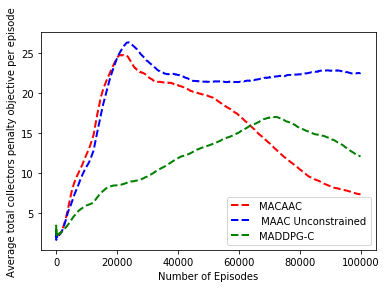}
                \caption{Expected total penalty of collectors}
                \label{pen-ctc-collec}
        \end{subfigure}%
        \begin{subfigure}{0.3\textwidth}
        \centering
                \includegraphics[width=0.95\columnwidth]{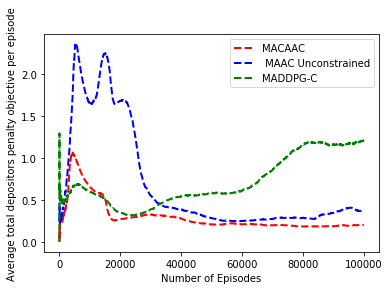}
                \caption{Expected total penalty of depositors}
                \label{pen-ctc-depo}
        \end{subfigure}%
        \caption{Performance of Algorithms on Constrained Cooperative Treasure Collection during the training. }\label{ctc-plots}
        \label{}
\end{figure}
\begin{table}[ht]
\centering
\renewcommand{\arraystretch}{1.3}
\begin{tabular}{|c|c|c|c|}
\hline
\begin{tabular}[c]{@{}c@{}}Name of \\ the Algorithm\end{tabular} & \begin{tabular}[c]{@{}c@{}}Average total cost \\ over $10,000$ iterations\end{tabular} & \begin{tabular}[c]{@{}c@{}}Average total penalty of collectors\\ over $10,000$ iterations\end{tabular} & \begin{tabular}[c]{@{}c@{}}Average total penalty of depositors\\ over $10,000$ iterations\end{tabular} \\ \hline
\textbf{MACAAC}                            & \textbf{-76.21}                                                                                    & \textbf{4.70}                                                                                                          & \textbf{0.15}                                                                                                        \\ \hline
MADDPG-C                                                           & -22.20                                                                                 & 7.59                                                                                                         & 0.76                                                                                                          \\ \hline
Unconstrained                                                    & -88.41                                                                                   & 13.99                                                                                                         & 0.35                                                                                                        \\ \hline
MACAAC with Fixed Weights                                                    & -54.68                                                                                  & 2.63                                                                                                         & 0.18                                                                                                        \\ \hline
\end{tabular}
\vspace*{0.3 cm}
\caption{Performance comparison of algorithms in testing phase on Constrained Cooperative Treasure Collection with penalty threshold for collectors ($\alpha1$) set to $12$ and that of the depositors ($\alpha2$) set to $0.2$. }
\label{t2}
\end{table}
\begin{figure}[ht]
\centering
        \begin{subfigure}{0.45\textwidth}
        \centering
                \includegraphics[width=0.95\columnwidth]{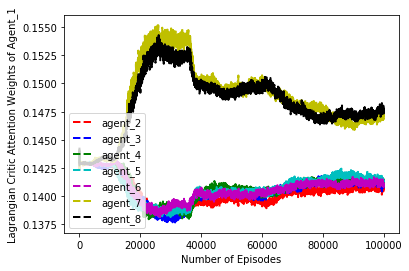}
                \caption{Attention Weights of Lagrangian critic}
                \label{MC-attention-1}
        \end{subfigure}%
        \begin{subfigure}{0.45\textwidth}
        \centering
                \includegraphics[width=0.95\columnwidth]{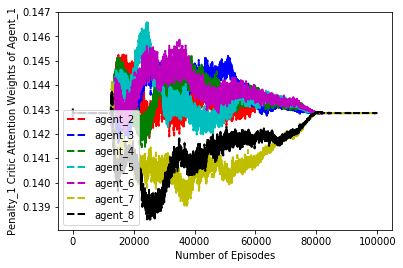}
                \caption{Attention weights of penalty$\_$1 critic}
                \label{Coll-attn-1}
        \end{subfigure}%
        \caption{Attention weights of Agent 1. These plots indicate the attention weights assigned to other agents by Agent 1}\label{ctc-attention-1}
\end{figure}

\begin{figure}[ht]
\centering
        \begin{subfigure}{.45\textwidth}
            \centering
            \includegraphics[width=0.95\columnwidth]{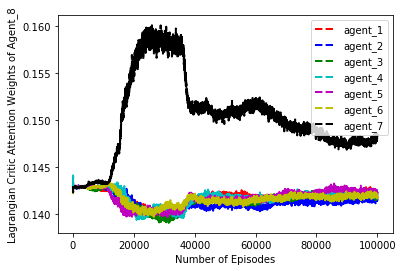}
                \caption{Attention Weights of Lagrangian critic}
                \label{MC-attention-8}
        \end{subfigure}
        \begin{subfigure}{0.45\textwidth}
        \centering
                \includegraphics[width=0.95\columnwidth]{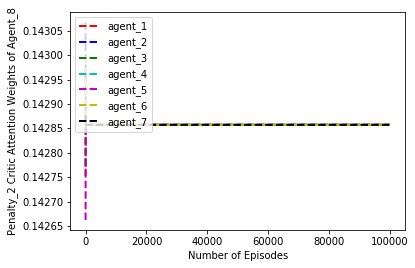}
                \caption{Attention weights for penalty$\_$2 critic}
                \label{Depo-attn-8}
        \end{subfigure}%
        \caption{Attention weights of Agent 8. These plots indicate the attention weights assigned to other agents by Agent 8}\label{ctc-attention-8}
\end{figure}

\subsection{Constrained Cooperative Treasure Collection}
\subsubsection{Description of the experiment}
In this experiment, we have a total of $8$ agents, out of which 6 agents are `collectors' (Agents $1,\ldots,6$, and the other two are `deposits (or banks)' (Agents 7 and 8). 
The role of collectors is to collect the `treasures' that are randomly generated in the environment and deposit them into the `banks' of the same color as the treasure. New treasures will be re-generated once the existing treasures are collected. The role of the `depositors' is to stay close to the collectors carrying their treasures. The length of each episode is 100 time-steps where all agents receive the shared single-stage cost associated with the total distances from their goals. Moreover, a cost of $-5$ (a positive reinforcement) is added every time a treasure is collected and deposited\footnote{This is the reason the costs in Table \ref{t2} are negative, as the agents learn to collect and deposit treasures}. We consider two penalty constraints in this experiment for collectors and depositors separately to demonstrate the effect of attention weights (discussed in Section \ref{attn-disc}). The penalty threshold for collectors ($\alpha1$) is set to $12$ and depositors ($\alpha2$) is set to $0.2$ and the discount factor is $0.99$.

\subsubsection{Discussion}
\begin{itemize}
    \item As in our previous experiment, we can observe from Figure \ref{cost-ctc} that the average total cost converges for all three algorithms. 
    \item From Table \ref{t2}, we can see that our proposed `MACAAC' satisfies the penalty constraints of both `collectors' and `depositors'. Moreover, the average cost obtained by `MACAAC' is lower compared to the `MADDPG-C' algorithm. As the agents do not incur a penalty in `Unconstrained', its average cost is least among three algorithms. 
\end{itemize}
\subsubsection{Discussion of Attention graphs}\label{attn-disc}
We now discuss the attention weights learned by the agents during the training. We present the attention weights learnt by the agent 1, which is a `collector' in Figure \ref{ctc-attention-1} and agent 8, which is a `depositor' in Figure \ref{ctc-attention-8}. Recall that, there are six collectors (agents 1-6) and two depositors (agents 7 and 8). In this experiment, there are three critics that use attention. Two penalty critics, which we refer to as penalty$\_$1 and penalty$\_$2 critics, compute the expected penalty costs of collectors and depositors respectively. The feedback from these critics is used in improving the Lagrange parameters (Step 30 of Algorithm \ref{C-AC}). Then, there is the main critic whose feedback is used to improve the policy parameters (Step 28 of Algorithm \ref{C-AC}). Note that the penalty critics make use of only the penalty costs whereas the main critic (Lagrangian critic) makes use of Lagrangian cost that involves both main cost and penalty costs (Step 20 of Algorithm \ref{C-AC}). 

In Figure \ref{MC-attention-1}, we observe that the Lagrangian Critic of agent 1 focuses more on the depositors\footnote{as  higher attention weights are assigned to agents 7 and 8} throughout its training. The agent 1 required to deposit its collected treasures into depositors, to minimize its cost, and hence the Lagrangian critic attends more to information of the depositors. The attention graph of Penalty$\_$1 critic of agent 1 in Figure \ref{Coll-attn-1} is very interesting. At the beginning of training, agent 1, to avoid collisions, focuses more on the information of other collectors and less on the depositors. However, as the training progresses, all the agents learn to move towards the depositors to deposit their treasures. Hence, the information of the depositors becomes very relevant for agent 1 to avoid collisions. This can also be confirmed from Figure \ref{pen-ctc-collec} where the constraint ($\alpha = 12$) is being satisfied, towards the end, after $80k$ iterations. Therefore, we observe that, towards the end of the training, agent 1 attends to information of all the other agents equally. In this way, the attention mechanism enables the agents to dynamically select relevant information during the training.

In Figure \ref{ctc-attention-8}, we report the attention graphs of agent 8, which is a depositor. In Figure \ref{MC-attention-8}, we observe that agent 8 attends more to the information of other depositors, i.e., agent 7. We have seen earlier that the collectors attend more to the information of the depositors to deposit their treasures. Moreover, the Lagrangian cost is a combination of the main cost and penalty costs. Therefore, agent 8 has to move in directions that don't overlap with agent 7, thereby providing collectors enough space to safely (avoiding collisions) deposit their treasures. Finally, in Figure \ref{Depo-attn-8}, we see that penalty$\_2$ critic of agent 8 attends to information of all the agents uniformly throughout its training.
Similar to the penalty$\_$1 critic of agent 1, information of all the agents is equally important for the agent 8 to avoid collisions. 

In this way, our proposed algorithm provides a framework for multi-agents to learn suitable attentions for various sub-tasks. The advantage of this paradigm can be seen from our results, where our proposed algorithm `MACAAC' performs well while satisfying the specified penalty constraints.



\section{Effect of Fixed weights for Constrained Costs}\label{ab-studies}
As discussed in the introduction section, a constrained problem could be solved by adding the constrained costs to the main cost. However, the weights to be associated with the constrained costs to satisfy the specified constraints are not known. Therefore, in our proposed algorithm, on a slower time-scale, Lagrangian parameters are iteratively learnt, which act as weights for the constrained costs. In this section, we investigate the effect of using constant and fixed weights for the constrained costs during the training. That is, we construct a cost function as $k + \sum_{j=1}^{m}w_jc_j$, where $k$ is the main cost function, $c_j, ~ 1 \leq j \leq m$ are $m$ constrained costs and $w_j$ is the weight associated with constraint $j$. The weights we use in the experiments are the converged Lagrange parameters from the ``MACAAC'' algorithm. We call this experiment ``MACAAC with Fixed Weights''. 

In ``Constrained Cooperative Navigation,'' there is one constraint and the weight assigned to this constraint is $w_1 = 5.534$. We observe from Table \ref{t1} that, this value of $w_1$ satisfies the penalty constraint $\alpha =3$. Moreover, the average cost obtained is slightly less than that of the standard ``MACAAC''. 

In Table \ref{t2}, we run the ``MACAAC with Fixed Weights'' for the ``Constrained Treasure Collection'' experiment. The weights assigned to two penalty constraints in this experiment are $w_1 = 3.47$ and $w_2 = 0.83$. Similar to our earlier experiment, we observe that these weights satisfy the penalty constraints of both collectors and depositors. However, the average cost obtained is higher than the standard ``MACAAC'' algorithm as these fixed weights may be too restrictive in this experiment. On the other hand, our proposed algorithm adaptively trains (increases or decreases) the Lagrange parameters during the training, leading to a better policy.   

From this study, we conclude the following:
\begin{enumerate}
    \item Our proposed ``MACAAC'' adaptively computes the Lagrange parameters that satisfy the penalty constraints.  
    \item Our proposed ``MACAAC'' algorithm computes a near-optimal solution using a two-time scale approach, where policy is updated on a faster timescale and Lagrange parameters are updated on a slower timescale. 
\end{enumerate}

\section{Conclusions}
\label{concl}
We have considered a constrained multi-agent RL setting where the agents need to learn optimal actions that satisfy the constraints specified on their actions. We have proposed an attention mechanism based constrained Actor-Critic algorithm that computes the Lagrange parameters on a slower time-scale and optimal policy on a faster time-scale. The attention mechanism enables the agents to select relevant information during the training, for computing the policy and satisfying the constraints. Through experiments on two benchmark multi-agent settings, we have shown that our proposed algorithm computes a near-optimal solution satisfying the penalty constraints. 

\section{Acknowledgements}
Raghuram Bharadwaj was supported by a fellowship grant from the Centre for Networked Intelligence (a Cisco CSR initiative) of the Indian Institute of Science, Bangalore. This work was supported by the Robert Bosch Centre for Cyber-Physical Systems, Indian Institute of Science, and a grant from the Department of Science and Technology, India. S.Bhatnagar was also supported by the J.C.Bose Fellowship.
\bibliographystyle{plain}
\bibliography{references}


\end{document}